\documentclass{article} % For LaTeX2e
\usepackage{iclr2025_conference,times}

\usepackage{amsmath,amsfonts,bm}

\def\eqref#1{equation~\ref{#1}}
\def\1{\bm{1}}

\DeclareMathAlphabet{\mathsfit}{\encodingdefault}{\sfdefault}{m}{sl}
\SetMathAlphabet{\mathsfit}{bold}{\encodingdefault}{\sfdefault}{bx}{n}

\usepackage{hyperref}
\usepackage{url}

\DeclareMathAlphabet{\fol}{OT1}{lmtt}{b}{n}
\usepackage{multirow}
\usepackage{makecell}
\usepackage{algorithm}
\usepackage{algorithmic}
\usepackage{listings}

\usepackage{booktabs}
\newcommand{\C}{\mathcal{C}}
\renewcommand{\P}{\mathbb{P}}
\newcommand{\F}{\mathcal{F}}
\renewcommand{\H}{\mathcal{H}}
\renewcommand{\S}{\mathcal{S}}
\newcommand{\Q}{\mathcal{Q}}
\newcommand{\D}{\mathcal{D}}

\newcommand{\T}{\mathbb{R}}
\newcommand{\B}{\mathbb{E}}
\renewcommand{\epsilon}{\varepsilon}

\usepackage{newfloat}
\usepackage{listings}
\floatstyle{ruled}
\newfloat{listing}{tb}{lst}{}
\floatname{listing}{Listing}

\title{An Empirical Study of Conformal Prediction in LLM with ASP Scaffolds for Robust Reasoning}

\author{Navdeep Kaur \& Lachlan McPheat \\
The Alan Turing Institute\\
London, UK\\
\texttt{\{nkaur, lmcpheat\}@turing.ac.uk} \\
\And
Alessandra Russo \\
Imperial College London \\ The Alan Turing Institute\\
London, UK\\
\texttt{\{arusso\}@turing.ac.uk}\\
\And
Anthony G. Cohn \\
The University of Leeds \\ The Alan Turing Institute\\
Leeds \& London, UK\\
\texttt{\{acohn\}@turing.ac.uk}\\
\And
Pranava Madhyastha \\
City University of London \\ The Alan Turing Institute\\
London, UK\\
\texttt{\{pmadhyastha\}@turing.ac.uk}\\
}

\iclrfinalcopy % Uncomment for camera-ready version, but NOT for submission.
\begin{document}

\maketitle

\begin{abstract}
%    The deductive abilities of Large Language Models (LLMs) alone leave something to be desired when it comes to generalisability and robustness. Building scaffolds surrounding LLMs in reasoning tasks is one method which promises to improve in these areas. In this paper, a scaffold is defined which forces LLMs to produce sets of Answer-Set Programs (ASPs) using prompting and conformal prediction techniques, improving robustness and accuracy on a spatial reasoning task containing a wide range of lengths of reasoning chains.
In this paper, we examine the use of Conformal Language Modelling (CLM) alongside Answer Set Programming (ASP) to enhance the performance of standard open-weight LLMs on complex multi-step reasoning tasks. Using the StepGame dataset, which requires spatial reasoning, we apply CLM to generate sets of ASP programs from an LLM, providing statistical guarantees on the correctness of the outputs. %An admission function based on ASP syntax filters valid samples, and evaluation metrics—including ROUGE-L and a new \textit{LLM-as-Judge} metric—measure the quality and diversity of the generated ASPs. 
Experimental results show that CLM significantly outperforms baseline models that use standard sampling methods, achieving substantial accuracy improvements across different levels of reasoning complexity. Additionally, the \textit{LLM-as-Judge} metric enhances CLM's performance, especially in assessing structurally and logically correct ASP outputs. However, calibrating CLM with diverse calibration sets did not improve generalisability for tasks requiring many more reasoning steps, indicating limitations in handling more complex tasks. \footnote{We thank Microsoft Research - Accelerating Foundation Models Research program, for the provision of Azure resources to access OpenAI models.  AGC also acknowledges partial support from the Economic and Social Research Council (ESRC) under grant ES/W003473/1.}
\end{abstract}

\section{Introduction}

%ICLR requires electronic submissions, processed by
%\url{https://openreview.net/}. See ICLR's website for more instructions.

%If your paper is ultimately accepted, the statement {\tt
%  {\textbackslash}iclrfinalcopy} should be inserted to adjust the
%format to the camera ready requirements.

%The format for the submissions is a variant of the NeurIPS format.
%Please read carefully the instructions below, and follow them
%faithfully.

Building automated reasoning systems is one of the cornerstone goals of research in artificial intelligence, with wide-reaching applications in a variety of domains. LLMs have demonstrated remarkable capabilities in various natural language processing tasks \cite{Min2021Recent}, yet exhibit critical limitations in a variety of reasoning-related tasks, particularly in multistep and spatially complex scenarios \cite{yang_coupling_2023,li_advancing_2024}. Although these models process and generate natural language with unprecedented fluency, their reasoning capabilities are still limited by several key constraints.

Most LLMs frequently struggle with verifiable, robust and interpretable reasoning. Experimental studies have shown significant variability in LLM responses to identical reasoning related questions which highlights inherent instability \cite{mirzaee2023disentangling}. For instance, sampling from an LLM's responses to reasoning-based queries can produce widely varying answers with significantly divergent reasoning traces. This lack of robustness undermines their reliability in critical decision-making contexts \cite{ji2023survey}. Moreover, tasks that require multiple sequential reasoning steps, such as complex spatial reasoning, often expose fundamental weaknesses in LLM's systematic reasoning capabilities \cite{shi_stepgame_2022}. 

Recent research has increasingly focused on neural-symbolic approaches that combine the strengths of LLMs in learning from data and rigorous reasoning capabilities of symbolic systems \cite{mao2019neural,yu2024natural}. Specifically, Answer Set Programming (ASP), a type of  declarative logic programming, has emerged as a promising candidate for enhancing LLMs' reasoning capabilities \citep{yang_coupling_2023}. Recent studies have explored integrating LLMs with ASP to transform natural language into logical representations, particularly for spatial reasoning tasks \citep{yang_coupling_2023,li_advancing_2024}.

However, existing approaches face significant challenges in generating reliable intermediate scaffolds: the logical representations that bridge natural language inputs and systematic symbolic reasoning systems. The translation of natural language into formal logic programs introduces substantial uncertainty, as LLMs can produce inconsistent or semantically incorrect translations \cite{li_advancing_2024}. This variability may become a significant weakness in this reasoning pipeline, potentially compromising the reliability of downstream reasoning tasks \cite{mirzaee2023disentangling}.

In this paper, we present an approach that exploits the recently proposed \textit{Conformal Language Modelling} (CLM) \cite{quach_conformal_2024} to systematically address these scaffold generation challenges. Conformal prediction (CP), the foundational framework for CLM, offers unique advantages in uncertainty quantification \cite{vovkAlgorithmicLearningRandom2005,angelopoulos_learn_2022}. 
%Unlike traditional sampling approaches, CP provides rigorous statistical guarantees with minimal assumptions, making it particularly suited for handling complex cases which are prominent in neuro-symbolic approaches. 
CLM has been applied to natural language generation tasks like summarisation and question-answering \cite{quach_conformal_2024}. Although recent advances in conformal prediction for language generation show promise, significant questions remain regarding the scalability, generalisability and robustness of these approaches across reasoning domains \cite{campos2024conformal}. In this paper, we present a thorough analysis of these open challenges through an empirical lens. 
We assess how calibration in CP affects generalisability on multi-hop reasoning tasks and evaluate the efficacy of adapting metrics for formal language generation (ASP). Additionally, we identify common error types in the ASP scaffolded LLM based reasoning pipelines and examine how these propagate. Our findings demonstrate the reliability and performance of neuro-symbolic reasoning systems, contributing to the development of more robust and interpretable  reasoning frameworks.

\section{Background}
\label{sec:bg}
    In this section we introduce the main components used in our work: conformal prediction, conformal language modelling and answer set programming.
    
    Conformal Prediction is a statistical framework introduced by \cite{vovkAlgorithmicLearningRandom2005} that provides a method for constructing prediction regions with finite-sample guarantees under minimal assumptions. CP can be described as a sampling method which, under certain mild assumptions, guarantees that samples taken from any distribution using this method will be `correct' with a controlled probability. %This theory is well-received in machine learning, where providing guarantees on sampling-correctness helps improve performance, as inference using neural networks is sampling from a theoretically unknown distribution. \textcolor{magenta}{CP has already proven its efficacy in image classification tasks \cite{???} drug discovery \cite{cortes-cirianoConceptsApplicationsConformal2019}}, for instance. 
    This theory has been well-received in machine learning, where providing guarantees on the correctness of predictions helps improve performance, in particular since inference using neural networks involves sampling from theoretically unknown distributions \cite{campos2024conformal}. CP has already proven its utility and efficacy in various domains, such as in computer vision tasks \cite{hechtlinger2018cautious} and drug discovery \cite{cortes-cirianoConceptsApplicationsConformal2019}.

    Conformal Language Modelling (CLM), developed in \citet{quach_conformal_2024}, is a recent generalisation of CP which allows conformal sampling of language models. CLM proposes a conformal approach to generating sets of responses with language models, allowing for statistical guarantees in the vast and complex output space of natural language generation. The main contribution of CLM is an algorithm that can control a specific risk: the expected indicator loss that the generated set contains at least one \textit{acceptable} response according to a predefined admission function—while simultaneously maximising other quantities of interest such as quality, diversity, and confidence. %This is no mean feat, as CP previously saw its main application in machine learning tasks with a far smaller output space than in language modelling. 
    However, it remains unclear how CLM affects generalisability, particularly because it has been primarily tested on natural language generation tasks like summarisation, and report generation, where verifying the accuracy of models is not always straightforward \cite{campos2024conformal}.
    %which is not helped by the types of tasks it has been experimented on in \cite{quach_conformal_2024} (summarisation, question-answering and report-generation) where there is not always a clear way to verify the accuracy of models. 
    %Applying CLM to domains with rigorous notions of correctness will tell us in better detail how it handles generalisability.
    Applying CLM to domains with rigorous notions of correctness, such as formal language generation, can provide better insights into its handling of generalisability and potential limitations.
    %In section \ref{sec:clm} we introduce the mechanics of CLM in further detail and subsequently adapt it to formal a language generation setting. 

    Answer set programming is a type of declarative programming which is commonly used in formal representations of knowledge, constraint satisfaction tasks, and formal common-sense reasoning. Answer set programs consist of finite sets of facts and rules that are used to generate \emph{answer sets} under \emph{stable model semantics}, which allows for efficient encoding of complex problems, notably in domains like graph search and combinatorial optimization~\cite{Lifschitz19}. ASP's utility extends to various industrial applications due to its expressive power and computational efficiency, for instance, in e-tourism \cite{riccaLogicBasedSystemETourism2010}, high-performance computing scheduling \cite{gamblinUsingAnswerSet2022}, configuration management, diagnosis, and planning \cite{erdem2016applications}.
    %ASP has been employed to personalise travel planning by reasoning over user preferences and constraints \cite{riccaLogicBasedSystemETourism2010}.
    %In high-performance computing scheduling, ASP aids in dependency checking to optimise job execution sequences \cite{gamblinUsingAnswerSet2022}.
    %Additionally, ASP has been instrumental in configuration management, diagnosis, and planning tasks where complex logical relationships need to be navigated effectively \cite{erdem2016applications}. 
    %The integration of ASP with neural approaches has also been explored to enhance reasoning capabilities in AI systems.
    Further, ASP has been used to bolster spatial reasoning in LLMs through neural-symbolic integration \cite{yang_coupling_2023}. 

    In the following section, we present a more formal introduction to Conformal Language Modelling (Section \ref{sec:clm}), which we then adapt to our purposes in Section \ref{sec:mtd}. With the adapted CLM, we set up an experiment in Section \ref{sec:exp_setup} executed on a reasoning task in Section \ref{sec:experiments} where we present and analyse the results. The main conclusions are presented in section \ref{sec:conclusion} where we also discuss future directions.
    
    %This approach has been extensively researched in both academia and industry, leading to applications in various domains like dependency checking in high-performance computing scheduling \cite{gamblinUsingAnswerSet2022}, e-tourism \cite{riccaLogicBasedSystemETourism2010}, and enhancing spatial reasoning in large language models (LLMs) through neural-symbolic integration \cite{yang2023coupling}.
    %The initial objects of study in answer set programming are answer set programs (ASPs) which are finite sets of facts and rules, which are used to generate answer sets in what is known as stable answer semantics which is a subset of the models of the ASP (viewed in the classical sense in logic programming). ASPs and their semantics are able to efficiently encode complex problems, the prototypical example being graph search. This has been researched in both academia and industrial settings, and is now used in various industrial applications \cite{dodaroCombiningAnswerSet2016} such as e-tourism \cite{riccaLogicBasedSystemETourism2010},dependency checking in HPC computing scheduling \cite{gamblinUsingAnswerSet2022} and \textcolor{magenta}{ ??? to name a few.}

\section{Formal introduction to CLM}
\label{sec:clm}
%% PRanava - I am editing directly inside here
    As mentioned in Section \ref{sec:bg}, CLM extends the statistical guarantees of CP to the domain of language modelling \cite{quach_conformal_2024}. CP provides statistical guarantees of sampling procedures while being agnostic to the underlying distribution, making its adaptation to LLMs a significant feat. We briefly summarise the mechanics of CLM below, setting the stage for the adaptations we make for \textit{formal} language modelling  in our experiments.
    
    At its core, CLM functions as a calibrated rejection sampling algorithm, allowing one to form \textbf{conformal sets of samples}, $\C(X)$, from an LLM, given a prompt $X$, such that the expected loss of $\C(X)$ is guaranteed to be within a chosen bound. 
    Concretely, for a given loss function $L$ and error tolerance $\delta \in (0,1)$, CLM ensures there exists a risk tolerance $\epsilon \in (0,1)$ such that for any prompt $X$, the conformal set $\mathcal{C}(X)$ satisfies the Equation \ref{eqn:coverage_theorem} with probability at least $1 - \delta$:
    %Explicitly, this can be stated using a loss function of your choice, $L$, and a chosen error-tolerance $\delta \in (0,1)$, there is an risk-tolerance $\epsilon\in(0,1)$ such that for any a prompt $X$, its conformal set of samples $\C(X)$ will have an expected loss bounded below by $\epsilon$ with probability $1-\delta$, stated as the following inequality: %$\E[L(\C(X))]\leq \epsilon$ with probability at least $1-\delta$ \cite{quach_conformal_2024}. In a single line this is 
    \begin{equation}\label{eqn:coverage_theorem}
        \P(\B[L(\C(X))]\leq \epsilon) \geq 1-\delta.
    \end{equation}
    
    This leads to the important question: How is the conformal set $\mathcal{C}(X)$ defined? In \cite{quach_conformal_2024}, the authors define $\mathcal{C}(X)$ using a rejection sampling algorithm that incorporates three key metrics: a) \textbf{quality of samples}: $\Q$, which measures how good each sample is; b) \textbf{diversity of samples}: $\S$, which ensures that the samples are sufficiently diverse; and c) \textbf{confidence of the set of samples}: $\F$, which evaluates the overall confidence in the correctness of the sample set. %Each of which is instantiated as metrics or measures. For instance $\Q$ measures the transition score of the final token of the sample $\Q(X,y) = -\log(p(y_{\mathrm{final}}|X))$ of the a greedy sample $y$, with final token $y_{\mathrm{final}}$, with respect to a prompt $X$ to the LLM. $\S$ measures the similarity of the samples of the LLM to ensure the conformal set has a sufficient diversity of entries. $\F$ measures a confidence of the set of samples being correct, which has been instantiated as the sum or maximum of confidences of the outputs in the set, each of which has been tested in \cite{quach_conformal_2024}. 
    
    Before considering these metrics however, one requires a method to filter out `unacceptable' samples. This is achieved using an \textbf{admission function}, $A$, which maps samples to $\{0,1\}$, effectively discarding samples that do not meet certain criteria. The choice of admission function can be tailored to specific tasks while still maintaining the coverage from Equation \ref{eqn:coverage_theorem}. Indeed, in \citet{quach_conformal_2024} we are presented with a range of instances of $A$. For instance,for the summarisation task $\mathrm{ROUGE}_L$ \cite{lin_rouge_2004} scores were used, which focuses on measuring the longest common subsequence between two samples. This is used to compare sample summaries $y$ to the references $X$ being summarised (concretely, $A(y) =1$ if $ \mathrm{ROUGE}_L(X,y)>0.35$ and $0$ otherwise).

    The construction of a conformal set $\mathcal{C}(X)$ in the context of language modelling given a prompt $X$ is done recursively using the chosen metrics, $\Q,\S,\F$ along with associated constraints $\lambda = (\lambda_1,\lambda_2,\lambda_3)\in\T^3$, which we learn in advance from the calibration process, explained in the following section. 
    
    The recursive construction of $\mathcal{C}(X)$ involves repeatedly sampling the LLM and rejecting samples which do not meet the constraints established by the admission function and the quality, diversity and confidence measurements $\Q,\S,\F$ introduced above. Concretely, we define a chain of sets $\C_0(X)\subseteq\C_1(X)\subseteq \cdots \subseteq \C_k(X)$, where $k$ is the sampling budget. For each $i=1,\ldots, k$ the set $\C_i(X)$ is defined according to Algorithm 1 in \citet{quach_conformal_2024}, which we simplify below. We will take $y_i$ to denote the $i^{th}$ sample of the LLM with prompt $X$, given that $y_i$ is admissible, i.e. $A(y_i)=1$\footnote{
        Note that it is indeed possible for none of the samples to be admissible, i.e. $A(y_i)=0$ for all samples $y_i$. This simply means the conformal set is empty.
    }. The initial set, assuming $A(y_0)=1$, is $\C_0(X)=\{y_0\}$. For each $i = 1,\ldots, k$ the set $\C_i(X)$ is defined as $\C_{i-1}\cup \{y_i\}$ if the  two conditions below are met:
    \begin{enumerate}
        \item $y_i$ is of high enough quality, i.e. $\Q(y_i|X)\geq \lambda_1$ 
        \item $y_i$ is diverse enough, i.e. $\S(y_i,y_j)\leq \lambda_2$ for  $0 \leq j < i$.
    \end{enumerate}
    Finally, after testing quality and diversity, we test whether the confidence of the set is sufficient: $\F(\C_i(X))\geq \lambda_3$. If this is true, we break the loop and return $\C_i(X)$ as the conformal set. If not, we repeat the loop for the next set $\C_{i+1}(X)$. 
    %% HOW IS THIS ONE?%%%
%The recursive construction a) starst with an empty conformal set. If the first sample $y_0$ is admissible ($A(y_0) = 1$), set $\mathcal{C}_0(X) = {y_0}$. Then, b) for each subsequent sample $y_i$ (where $i = 1, \ldots, k$, and $k$ is the sampling budget):
%verify that the sample meets the quality threshold: $\mathcal{Q}(y_i | X) \geq \lambda_1$.
%and also verify that the sample is sufficiently different from all previously accepted samples: $\mathcal{S}(y_i, y_j) \leq \lambda_2$ for all $j < i$. %Then, c) If both checks pass, add $y_i$ to the conformal set: $\mathcal{C}i(X) = \mathcal{C}{i-1}(X) \cup {y_i}$.Finally, after adding a sample, verify whether the confidence metric meets the threshold: $\mathcal{F}(\mathcal{C}_i(X)) \geq \lambda_3$. If it does, stop and return $\mathcal{C}_i(X)$ as the conformal set. If not, continue to the next sample.  
    Next, we present the calibration procedure of \cite{quach_conformal_2024} to explain how to find appropriate constraints $\lambda$ in the first place.

    \subsection{Calibration}
    First of all one identifies a calibration set $\D_{\mathrm{cal}}$, often a subset of validation or training data. This data is used to calibrate the values of $\lambda$ to ensure one's samples conform to a desired standard. The calibration procedure in \citet{quach_conformal_2024} is an adaptation of the Learn Then Test (LTT) procedure of \citet{angelopoulos_learn_2022}. %We briefly summarise the calibration procedure below, and refer to \citet{quach_conformal_2024} for further detail. 
    The calibration procedure is basically a search over a space of weighted configurations $\lambda = (\lambda_1,\lambda_2,\lambda_3)$\footnote{
        In general, there is a value of $\lambda$ for each metric, meaning that the configuration-tuple $\lambda$ can be of arbitrary length.}.
    Given the set of all possible configurations $\Lambda = \{0, 0.01, \ldots, 0.99, 1\}^3$, we consider for each $\lambda \in \Lambda$ and each possible risk-tolerance $\epsilon \in \{0,0.01,0.02,\ldots,0.99, 1\}$ one calculates the binomial tail bound $p$-value under the null-hypothesis $\H_\lambda: \B[L(\lambda)] > \epsilon$. This gives us a $p$-value $p_{\lambda,\epsilon} = p_\lambda$:
        \[p_\lambda := \mathbb{P}(\mathrm{Bin}(n,\varepsilon)\leq n\hat{R}_n(\lambda)),\] 
    where $\hat{R}_n(\lambda)$ is the \textit{empirical risk}, taken to be: 
        \[\hat{R}_n(\lambda) := \frac{1}{n}\sum_{i=1}^nL_i(\lambda)\]
    and $L_i(\lambda)$ is the \textit{loss}: 
        \[L_i(\lambda) := \mathbf{1}\{\not\exists y\in \mathcal{C}_\lambda(X_i) \mid A(y) = 1\}.\]
    Following this lengthy sequence of calculations of $p$-values, one applies a family-wise error rate controlling algorithm to the set of $p$-values to return the set of valid parameters $\Lambda_{\mathrm{valid}}$. In \citet{quach_conformal_2024} this is implemented using the `Bonferroni correction': $$\Lambda_{\mathrm{valid}} = \left\{\lambda \mid p_\lambda \leq \frac{\delta}{|\Lambda|}\right\}.$$ 
    
    Finally, we identify the configuration $\lambda \in \Lambda_{\mathrm{valid}}$ that minimises the below combination of conformal set size and number of samples needed to predict the correct label over the calibration set $\D_{\mathrm{cal}}$:$$%\hat{\lambda} := \mathrm{argmin}_{\lambda\in\Lambda_{\mathrm{valid}}} 
    \frac{1}{|\D_{\mathrm{cal}}|}\sum_{(X,Y)\in\D_{\mathrm{cal}}}\left(\rho_1|\mathcal{C}_\lambda(X)| + \rho_2\frac{[S_\lambda(X)-S^*(X)]^+}{S_\lambda(X)} \right)$$ where $S_\lambda(X_i)$ is the total number of samples made before rejecting (e.g. 20), $S^*(X_i)$ is the index of the first correct sample, and $[\cdot]^+=\mathrm{max}(\cdot, 0)$. The numerator $[S_\lambda(X)-S^*(X)]^+$ quantifies how quickly the LLM generates a correct sample. Since samples are generated sequentially, we ideally want to minimise superfluous generations which is done by minimising this number. We take $\rho_1,\rho_2$ to be $0.5$, as done by \citet{quach_conformal_2024} where the reader can find further detail.

\section{Methodology}
\label{sec:mtd}
In this work, we are specifically interested in exploring the application of CLM to improve the performance of mid-range open-weight language models of modest size on complex reasoning tasks. Specifically, we focus on the StepGame dataset of \cite{li_advancing_2024}, which is a correction
%\footnote{We refer the reader to \cite{li_advancing_2024} for an analysis of these corrections.} 
of the original dataset developed in \cite{shi_stepgame_2022}, which presents spatial reasoning challenges requiring multiple steps of logical deduction. Unlike previous approaches that leverage larger, more computationally intensive LLMs with extensive prompt engineering, our methodology emphasises improving the capabilities of a mid-range LLM through the integration of ASP and conformal prediction techniques. The goal here is to empirically validate the utility of CLM especially for ensuring that the generated ASPs indeed maintain highly useful outputs with highly accurate answer sets and to carefully conduct experiments investigating the generalisability across different calibration settings. 

The core construct of ASP is the logic rule, formulated as:     
\begin{equation*}
a_{1} \mid \ldots \mid a_{n} \; :- \; b_{1}, \ldots, b_{k} \; ; \; \mathbf{not} \; b_{k+1}, \ldots, \; \mathbf{not} \; b_{m}
\end{equation*}
Here, $a_i$ and $b_i$ are \textit{atoms} or positive \textit{literals}, and $\mathbf{not} \, b_j$ are negative \textit{literals}. Informally, this rule states that at least one of the $a_i$ must be true if all $b_i$ are true and all $b_j$ are false. A rule with no $b_i,\mathbf{not} \, b_j$ is called a \textit{fact}. ASP programs consist of sets of such rules and facts, which collectively define the problem space and constraints. ASP is particularly well-suited for multistep reasoning tasks due to its ability to model complex constraints and perform efficient search procedures to find solutions that satisfy all conditions. 
%This has been used for spatial reasoning tasks, in particular with the StepGame dataset in \cite{li_advancing_2024} where 99\% or 100\% accuracy was achieved by combining the Davinci LLM with ASP reasoning across all steps sizes. 

In our approach, CLM is employed to ASP programs sampled from the LLM such that the generated programs are not only syntactically valid but also, the answer sets are likely to contain the correct solution when processed by the \texttt{clingo} ASP solver \cite{gebser2018multishotaspsolvingclingo}. 
%A critical component of CLM is the calibration set, which consists of previously solved examples used to adjust the confidence levels and improve the generalisability of the model.
We investigate how different types of calibration sets—ranging from those containing single-step reasoning examples to those with multiple-step reasoning examples—affect the performance and generalisability of CLM. This focus allows us to understand the importance of sample diversity in the calibration phase and its impact on the model's ability to generalise across varied reasoning tasks.

Our core methodology involves translating natural language inputs from the StepGame dataset into ASP programs using the LLM. To facilitate the translation of natural language task descriptions into Answer Set Programming (ASP) code, we employ \textit{In-Context Learning (ICL)}. ICL exploits the inherent ability of large language models (LLMs) to discern patterns and structures from provided examples within the input prompt, enabling them to generate contextually appropriate outputs without explicit parameter fine-tuning. The entire ASP program is then sampled using CLM, producing a conformal set of programs. These ASP programs are then processed by \texttt{clingo} to derive answer sets. 

Note that the CLM process itself is carried out in two main stages with the first stage associated with learning $\lambda$ using the calibration set (see Calibration in Section \ref{sec:clm}). After the initial calibration, we employ a separate validation set to fine-tune the parameters further. This validation set consists of additional StepGame tasks not included in the calibration phase, providing an unbiased dataset to evaluate the performance of different $\lambda$ configurations and optimal $\epsilon$ values. This helps us identify the optimal $\lambda$ and $\epsilon$ settings that maximise the accuracy and reliability of the generated ASP programs. Once the best $\lambda$ and $\epsilon$ values are determined, they are fixed and used to evaluate the model's performance on an independent test set, ensuring that the selected parameters generalise well to unseen data. We next provide the exact experimental setup and details of our empirical analysis. 

\section{Experimental setup}\label{sec:exp_setup}
    
\subsection{Dataset} 
    All our empirical results are based on the StepGame dataset. The StepGame dataset of \citet{li_advancing_2024}, %which builds on that of \citet{shi_stepgame_2022},
    consisting of story-query-answer tuples of the form $\langle d,q,a \rangle$ where $d$ is a list of natural language descriptions of edges of a graph. These edges are in one of nine configurations: \textit{right, top-right, top, top-left, left, bottom-left, bottom, bottom-right}, and \textit{overlap} which are present in $d$ using a variety of synonyms. $q$ is a natural language query of the relation between two nodes in the graph, and $a$ is the answer. 
    %An example datapoint is:
    %\begin{quote}
    %\textcolor{magenta}{Example of a StepGame entry}    
    %    \[\left\langle 
    %    \begin{array}{c}
    %         d:(A\text{ is to the west of }B,B\text{ is above }C), \\
     %        q:\text{what is the relation between }A \text{ and }C\text{?},\\
     %        a:\text{top-left}
    %    \end{array} 
    %    \right\rangle\]
    %\end{quote}
    The entries in $d$ differ in the number descriptions of the graph, between $1$ and $24$, corresponding to the number of hops. 
    %The example above is an instance of a $2$-hop entry. 
    
    %This dataset is used in question-answering research and was defined as an extension to the bAbI dataset's spatial reasoning tasks \cite{westonAICompleteQuestionAnswering2015}, increasing the number of reasoning steps, or hops, that are needed to answer the questions from a single hop in \citet{westonAICompleteQuestionAnswering2015} to a range of hop from $1$ to $24$.
    Stepgame dataset has been used to test the spatial reasoning abilities of LLMs, where the multiple steps of reasoning are necessary to induce the relation between the two given nodes in each query as the answer is not directly retrievable in the description alone. Further variants of this dataset are available in \cite{shi_stepgame_2022} where noise in several different forms is added to the descriptions in the test set. We do not consider these noisy variations here, but leave it for future work. 
    %In general, LLMs perform worse when the number of reasoning steps increase \cite{li_advancing_2024}.

    For our experiments we generated a dataset of 465,975 datapoints split into 341,284 for training, 11,350 for validation and 113,341 for testing. Within each of these sets there is a distribution of different numbers of hops. The generation was performed using the \texttt{asp-solution.py} script\footnote{
     https://github.com/Fangjun-Li/SpatialLM-StepGame/blob/main/asp\_solution.py
     } written for the data-generation in \cite{li_advancing_2024}.  
     %Full details of the samples we used to ensure replicability will be provided in an online Github page upon acceptance.  

     The first experiment was performed on $6\times 200 = 1200$ from the StepGame test set, where we randomly sampled 200 entries of $1$, $2$, $3$, $4$, $5$ and $15$ hops respectively. The second experiment only on the entries with $1$, $2$ and $3$ hops.

\subsection{Admission}
   We designed an admission function to filter out non-programs from the samples generated by the LLM, done using an ASP syntax-check. Formally, this admission function, $A_{\mathrm{syntax}}$ say, is defined on samples $y$ as: 
   $$A_{\mathrm{syntax}} = \begin{cases}
      1 & \text{if \texttt{clingo} parses }y\text{ as a valid program} \\
      0 & \text{if \texttt{clingo} cannot parse } y
   \end{cases}$$
   The theory proving the inequality in Equation \ref{eqn:coverage_theorem} requires the admission function to return values in $\{0,1\}$, leaving us the freedom to use $A_{\mathrm{syntax}}$. Enforcing syntactic correctness ensures that only syntactically valid ASPs are admitted into the conformal set. This offers a significant advantage over similarity based admission functions, such as ROUGE scores commonly used in NLG tasks like summarisation (used in \citet{quach_conformal_2024}). While ROUGE metrics measure the overlap between generated text and reference summaries, they do not guarantee the functional or structural integrity of the output. In contrast, syntactic correctness provides an objective and reliable estimate of a sample's utility.

\subsection{Metrics and measures}
    %As for the metrics, our first experiment kept them the same as in the text-summarisation task in \cite{quach_conformal_2024}. 
    In our first experiment, we adopted the same evaluation metrics used in the text summarisation study by \citet{quach_conformal_2024}. 
    Specifically, we used the quality metric $\Q$ to be the average transition score of the output and for the confidence function $\F$; we then take the maximum of the transition scores of the output of the LLM. For the diversity metric we used $\S = \mathrm{ROUGE}_L$. However, since $\mathrm{ROUGE}_L$ measures sentence-level similarity, it may not be ideal for our formal language-generation, as textually similar strings may still have formal syntactic differences giving rise to unrelated answer sets. Despite this limitation, we justify the adoption of these metrics due to the conceptual similarity between our task and summarisation, as both involve condensing information into a structured format, albeit in different languages.
    %We justify this choice of metrics by the similarity in the task we are asking of the LLM, viewing translating natural language into ASPs as a type of summary, albeit in a formal language and perform a second experiment with the following purpose-built metric. % In \cite{quach_conformal_2024} four different instances of the confidence function are presented. 

  %  \textcolor{magenta}{For our second experiment we implement a LLM-as-judge metric, inspired by ???. The LLM-as-judge metric approach relies on having access to a finetuned language model. In our case, we finetuned $\fol{llama3.1-8B-Instruct}$ on roughly $\frac{1}{3}$ of the training set, containing varying description lengths from 1 to 24, taking a representative number of each length into the finetuning. The finetuning itself was performed using the LoRa adapter from the llamafactory repository \cite{zheng_llamafactory_2024}. The implementation of the LLM-as-judge metric is simply to take the difference of the logprobs of the two strings with respect to the finetuned model.}

    For the second experiment, we implement an \textit{LLM-as-Judge} metric, inspired by recent methodologies that leverage language models for evaluating and comparing generated outputs \cite{liu2023g,zheng2023judging}.  This approach uses a fine-tuned language model to assess the quality of generated ASP samples by computing the difference in average log probabilities between pairs of samples. Specifically, we fine-tuned the \texttt{llama3.1-8B-Instruct} model on approximately one-third of the training set. This subset includes a representative distribution of reasoning steps ranging from 1 to 24, ensuring that the model is exposed to diverse reasoning complexities during fine-tuning. The fine-tuning process was conducted using the LoRa adapter from the LlamaFactory repository~\cite{zheng_llamafactory_2024}, allowing efficient adaptation of the model with minimal computational overhead. Notably, the fine-tuned model achieves 100\% accuracy on the task, ensuring reliable evaluation performance.  

    The implementation of the \textit{LLM-as-Judge} metric involves calculating the difference in average log probabilities assigned by the fine-tuned model to two ASP samples. Formally, taking the samples of the fine-tuned language model to be $(y_i)_{i=1,\ldots,n}$ and $(z_j)_{j=1,\ldots,m}$, the \textit{LLM-as-judge} metric is evaluated on the pair $((y_i),(z_j))$ as: 
    $$\frac{1}{n}\sum_{i=1}^n \log p(y_i|LLM) - \frac{1}{m}\sum_{j=1}^m \log p(z_j|LLM).$$
    %Provide a formal definition here … 

    %This metric functions similarly to the \textit{LLM-as-Judge} approach, wherein the language model serves as an evaluator that quantitatively distinguishes between higher and lower quality samples based on their likelihood under the model's learned distribution. This metric provides an assessment of each sample's plausibility and adherence to the task-specific patterns learned during fine-tuning.

\subsection{Calibration}
%    We calibrate on two different sets to test see how it affects the generalisability in CLM. The first calibration set consists of StepGame entries of description lengths 1 to 5, and the other one of length 1 only. Both calibration sets are taken from the validation set and consist of 500 entries each. The mixed one maintains the distribution of different description length examples as the test set.

    To evaluate the impact of calibration set composition on the generalisability of Conformal Language Modelling (CLM), we used two distinct calibration sets. The first calibration set comprises StepGame entries where the number of reasoning steps ranges from 1 to 5, encompassing a diverse array of reasoning complexities. The second calibration set is restricted to entries with a only 1 reasoning step. Both calibration sets were extracted from the validation subset and consist of 500 entries each, ensuring a balanced and sufficient sample size for reliable calibration.

    The mixed calibration set (1-5 hops) mirrors the distribution of number of reasoning steps present in the test set, thereby maintaining consistency between calibration and evaluation phases. We are interested here in assessing the constraints CLM has on its generalisation capabilities in different constrained scenarios. We are interested in determining the extent to which the diversity of calibration data influences the model's ability to generalise to more complex and varied reasoning tasks.
\subsection{In context learning}
    %We use two different ICL-prompts, each consisting of two examples of the StepGame dataset, presented in full in appendix \ref{app:ICL_prompts}. The first ICL-prompt contains examples of description length 2 and 4, whereas the other ICL-prompt contains two distinct examples with description length 1. We use the prompt with description lengths 2 and 4 when the calibration has been done on examples of description lengths 1-5. Similarly, we use the description length 1 prompt when the calibration has been done on description length 1 data. This distinction is repeated in the baseline experiments too, to give a fairer comparison to the CLM results.

    We designed two ICL prompts, each containing two exemplar pairs from the StepGame dataset. These prompts are comprehensively detailed in Appendix \ref{app:ICL_prompts}. The first ICL prompt includes examples with 2 and 4 reasoning steps respectively, representing intermediate levels of reasoning complexity. The second ICL prompt consists of two 1-hop examples, focusing on single-step reasoning tasks. The selection of prompts is directly aligned with the calibration sets: the prompt containing 2 and 4-hop examples is used when calibrating with the mixed set (1-5 hops), while the prompt with 1 reasoning step is employed when calibrating with the single-length set. This distinction is repeated in the baseline experiments too, to give a fairer comparison to the CLM results.

\subsection{Language models}
    For our experiments, we used Meta's \texttt{llama3.1-8b-Instruct} model \cite{grattafiori_llama_2024} as the foundational language model around which our CLM framework is constructed. This model represents the smallest yet highly performant open-weight language model available at the time of our study, achieving competitive results on standard language model benchmarks \cite{srivastava2022beyond}. The selection of \texttt{llama3.1-8b-Instruct} is primarily motivated by our focus on enhancing the capabilities of less complex models through CLM. Using a smaller model will help us test rigourously the capability of CLM in enhancing the reasoning capabilities of LLMs without relying on extensive parametrisation or specialised training procedures. 

\section{Experiments and results}\label{sec:experiments}
\label{sec:exp}
\label{sec:res}
    % We adapt the experimental setup of \cite{quach_conformal_2024} together with in-context-learning prompts to build conformal sets of ASPs on which we run the answer set solver, clingo \cite{gebser2018multishotaspsolvingclingo}. We measure correctness of our conformal sets in line with the CP-literature, where a set is considered `correct' if it contains a correct sample. In our setting a sample ASP is considered correct if it  generates a single answer set whose only answer predicate is the correct one, following the accuracy measure of \cite{yang_coupling_2023}.
    \subsection{Experiment 1}
        In this experiment, we evaluated the effectiveness of CLM in generating accurate ASPs for StepGame dataset. Specifically we employ the following list of metrics and functions througout this experiment.  We use confidence $\Q$ as in \citet{quach_conformal_2024}, $\mathrm{ROUGE}_L$ \cite{lin_rouge_2004} as our diversity-metric $\S$, and the maximum logprob over the set as our confidence function $\F$\footnote{That is $\F(\C(X)) = \max\{-log(p(y|X))\mid y\in \C(X)\}$}. We also instantiate the admission function $A$ using \texttt{clingo} to test the syntactic correctness of the sample\footnote{We employ some minor post-processing to the samples. We found the LLM tended to repeat lines of code which would be cut off by the max-token limit, causing syntactic errors. Hence we remove any incomplete lines of code produced by the LLM. In early experiments we identified a short list of tokens which caused syntactic errors, which we used as stopping-tokens at inference time.}.
    
        We tested the two different calibration settings on six portions of the test-set, split according to multi-step reasoning including 1, 2, 3, 4, 5 and 15-steps to test the efficacy of calibration on both in and out-of-distribution data. %The LLM used was Meta's \texttt{llama3.1-8B-Instruct} model \cite{grattafiori_llama_2024}.

   \begin{table}[h!]
        \centering
        \begin{tabular}{l l l l l l l}
        %\Xhline{3\arrayrulewidth}
        \toprule
        \textbf{\# hops $\rightarrow$ }   &   1&   2&   3&   4&   5&   15 \\
        \midrule
        $\fol{Calib.\,1-5}$    & 80.0  & 72.5 & 70.5 & 70.0 & 65.0 & 0.5 \\
        $\fol{Calib.\,1}$      & 71.5 & 65.5 & 55.5 & 54.0 & 50.0 & - \\
        \midrule
        $\fol{Basel.\, 2+4}$   & 45.0 & 39.5 & 36.5 & 30.0 & 28.5 & 0 \\ 
        $\fol{Basel.\, 1}$     & 42.0 & 28.5 & 27.5 & 19.5 & 13.5 & 0 \\ 
        \bottomrule
        \end{tabular}
        \caption{CLM results (accuracies in \%) on StepGame task, using the ASP-parser as the admission function. $\fol{Calib.}$ refers to \textit{Calibration Length}, where \textit{1-5} refers to calibration on 1 to 5-hop StepGame entries, and $1$ to calibration on single-hop entries. $\fol{Basel.}$ refers to \textit{Baseline} and $\fol{2+4}$ and $\fol{1}$ refer to the type of ICL prompt used, see Section \ref{sec:exp_setup} for further detail.}\label{tab:main_table}
        \end{table}

        The baseline experiments were instantiated using the same LLM, where we provided the same ICL prompts as for the CLM sampling but then only sampled once, greedily, maintaining all other hyperparameters. The LLM-output is post-processed and then fed to \texttt{clingo} to generate answer sets. Note that we consider a sample program correct 
        %if there is only a single answer set generated by \texttt{clingo} which contains only one answer-predicate which is the correct one.
        iff there is only a single answer set generated by \texttt{clingo} and it contains only one answer-predicate and that answer predicate  is the correct one.
        
        The results of this experiment are presented in Table \ref{tab:main_table}, where we observe firstly that CLM significantly improved accuracy across all test sets by at least 20 percentage points in comparison to the baseline results. The calibration set consisting of 1 to 5-hop examples (`$\fol{Calib.}$ $\fol{1}$ - $\fol{5}$') consistently outperformed the calibration set limited to single hop (`$\fol{Calib.}$ $\fol{1}$'). This trend is evident across all test segments, with differences becoming more pronounced as the number of reasoning steps increased. This highlights the importance of diverse calibration data in enabling the model to generalise effectively to more complex scenarios. 

        One notable aspect of the experiment is the evaluation on the 15-hop test set, which falls outside the range of both calibration settings. Here, the performance drops significantly and the generated answer sets are either null or mostly incorrect. This result underscores the inherent challenge of generalising to significantly longer sequences without specific calibration data. It suggests that, while CLM with diverse calibration improves generalisation, there are still considerable difficulties in handling sequences that require substantially more reasoning steps than those seen during calibration. Finally, the improvements in performance with CLM underscore the capability of ASP to effectively capture and reason through complex logical relationships.

    \subsection{Experiment 2}
        %Our second experiment was to use the purpose-built LLM-as a judge metric introduced in section \ref{sec:exp_setup}. The remaining metrics, admission function and hyperparameters were kept the same as in the first experiment. 
        In our second experiment, we evaluated the effectiveness of the purpose-built \textit{LLM-as-Judge} metric, as introduced in Section \ref{sec:exp_setup}. This new metric was designed to overcome some of the inherent limitations of conventional metrics, such as the ROUGE-L score, when applied to formal language tasks like ASP generation. The remaining metrics, admission functions, and hyperparameters were kept identical to those in Experiment 1 to allow for a direct comparison of performance outcomes. The results are presented in Table \ref{tab:llm_as_judge}.
        \begin{table}[h!]
        \centering
        \begin{tabular}{@{}lllll@{}}
            \toprule
            \textbf{\# hops $\rightarrow$ }            &                   & 1     & 2 & 3 \\ 
            \midrule
            \multirow{2}{*}{$\fol{Calib.\,1-5}$}   & $\fol{LLM}$ $\fol{as}$ $\fol{Judge}$      & 80.0  & 72.5 & 72.0 \\
                                                & $\fol{ROUGE-L}$ $\fol{score}$     & 80.0  & 72.5 & 70.5 \\
            \midrule
            \multirow{2}{*}{$\fol{Calib.}\,1$}     & $\fol{LLM}$ $\fol{as}$ $\fol{Judge}$      & 76.5  & 68.5 & 60.5 \\
                                                & $\fol{ROUGE-L}$ $\fol{score}$     & 71.5  & 65.5 & 55.5 \\ \bottomrule
        \end{tabular}
        \caption{Results (accuracies in \%) from using LLM-as-judge as a similarity metric in the CLM and calibration procedures.}\label{tab:llm_as_judge}
        %\vspace{-0.20in}
        \end{table}

        We observe that the \textit{LLM-as-Judge} metric provided a consistent improvement over the ROUGE-L score across the calibration settings, especially for the more complex descriptions (lengths 2 and 3). 
        %For the mixed calibration set (`$\fol{Calib.}$ $\fol{1}$ - $\fol{5}$'), the accuracy of the \textit{LLM-as-Judge} metric was 80.0\%, 72.5\%, and 72.0\% for descriptions of lengths 1, 2, and 3, respectively.
        This performance closely matches or slightly outperforms the $\mathrm{ROUGE}_L$ metric for each corresponding number of reasoning steps. %In particular, for the 3-hop test set, \textit{LLM-as-Judge} yielded an accuracy of 72.0\% compared to 70.5\% for the $\mathrm{ROUGE}_L$ score.
        This improvement suggests that \textit{LLM-as-Judge} is better at assessing the quality of complex reasoning samples by capturing the nuances of logical structure that are essential for ASP correctness. For single-length calibration set, \textit{LLM-as-Judge} metric demonstrated a marked improvement over the ROUGE-L score across all numbers of reasoning steps. 

        This experiment demonstrates the primary limitation of the ROUGE-L metric, as observed in these experiments, lies in its focus on n-gram similarity, which is less suitable for evaluating formal languages like ASP. ROUGE-L does not adequately capture structural or functional correctness, leading to the acceptance of syntactically incorrect or logically inconsistent samples.

\subsection{Error analysis}
We found there are five type of errors that are made by the LLM while generating answer set programs. We list them below in decreasing order of prevalence.
\noindent
The most prominent error is when \texttt{clingo} returns `$\fol{empty}$'. This is because \texttt{clingo} failed to execute the ASP program successfully. This might be because: (i) The LLM generated a wrong fact corresponding to an instruction; for example, for a given instruction `L is at F's 9 o' clock', the LLM may generate \lstinline{at("L", "F", 9)} which is 
%neither 
none
of the nine possible configurations in Section \ref{sec:exp_setup}. (ii) The rules have typos in them, for example: \lstinline{is(A, down_right, B):- down_right(B)} (iii) The LLM generates gibberish at the end of a correct ASP program %(e.g., it generates \lstinline{Example 4:} at the end of an otherwise correct program, to mimic the ICL prompt). 
This makes \texttt{clingo} raise a parse error 
%- e.g.   \lstinline{Example 4:} is not valid \texttt{clingo} syntax.) 
(iv) Instead of generating a valid ASP program,  the LLM generates natural language  text which is read as gibberish by \texttt{clingo}.

\noindent
The second most prominent kind of error is when \texttt{clingo} returns one answer set missing any answer\footnote{The answer in this case is a fact of the form $\fol{answer(}R\fol{}\fol{)}$, where $R$ is one of the nine relations.}. Such answer sets are not empty, as answer may contain facts it has been given or other inferred facts. This could be due to at least one of the ASP facts corresponding to an instruction being wrong.
%, such as:
%\begin{lstlisting}
% I is there and M is at the 2  position of a clock face.
%clock("M", 2).
%\end{lstlisting}
%\noindent
The third most common type of error is when \texttt{clingo} returns one answer set containing multiple answers. This could be due to wrong rule being generated by LLM such as:

\begin{lstlisting}
is(A, down_left, B):- down_left(A,B).
is(A, down_right, B):- down_left(A,B).
\end{lstlisting}

\noindent
When this rule is fired, it would lead to two answers for a given query.

\noindent
The fourth most common type of error is when \texttt{clingo} returns one answer set containing one answer but the answer is incorrect. The most common reason is that translation done by the LLM from NLP to ASP facts is wrong. For example:
\begin{lstlisting}
% U is positioned above N and to the right.
top("U", "N").
\end{lstlisting}
In the above instruction, the correct answer generated by the LLM should have been $\textit{top-right("U", "N")}.$

The final type of error is when \texttt{clingo} returns multiple answer sets for a given input instruction. In principle, there are plenty of reasons for ASP programs to have such answer sets, but in this case this is due to the LLM generating multiple facts corresponding to one NLP instruction in a single row separated by commas, for example:
\begin{lstlisting}
% F is on the right side and below X.
right("F", "X"), down("F", "X").
\end{lstlisting}
The comma in the two facts acts as disjunction, meaning that \texttt{clingo} may use either of the two facts, resulting in two answer sets for the same program.

\section{Conclusions}
\label{sec:conclusion}
%\subsection{Conclusions}
%     \begin{enumerate}
%         \item CLM in this setting does improve the accuracy on StepGame comparing bet baseline accuracies to worst CLM accuracies.
%         \item LLM-as-judge improves the CLM
%         \item Calibrating on a diverse set does not improve the generalisability to length 15 tasks...
%     \end{enumerate}
% \subsection{Outlook}
%     \begin{enumerate}
%         \item Need further investigation on the effect of calibration set size.
%         \item Study the statistics of conformal set size.
%         \item Try the same setup on other tasks such as bAbI, for more spatial reasoning, but also logic puzzles and planning tasks.
%         \item other notions of accuracy \textcolor{magenta}{this has been fleshed out below (repurposed from previous accuracies-section)}
%     \end{enumerate}

    In this study, we explored the application of CLM to enhance the performance of a mid-range open-weight language model on complex reasoning tasks, specifically using the StepGame dataset. The CLM integration significantly improved accuracy on the StepGame task. When compared to baseline models that relied on standard sampling methods, CLM consistently outperformed across all tested segments. We also observe that the \textit{LLM-as-Judge} metric further improved the performance of CLM, in particular on multi-step reasoning. We observed that the diversity of samples is an important consideration for the calibration set. However, we also observed that calibrating CLM on a diverse set of examples (1 to 5 reasoning steps) did not enhance generalisability to tasks requiring significantly more reasoning steps (e.g. 15-hop). These results underscore the potential of CLM in augmenting the reasoning capabilities of standard language models, particularly when complemented by advanced evaluation metrics like \textit{LLM-as-Judge}. However, the diminished performance on higher-complexity tasks shows the potential limitations of CLM. Future work could focus on a thorough examination of the statistics related to conformal set sizes, such as their distribution and relationship with task complexity. This could inform more effective sampling strategies and constraint settings within the CLM framework.  Repeating the experiments multiple times to enable the calculation of error bars would also be desirable. Finally, a further baseline could be provided consisting of the LLM alone, without recourse to the external ASP reasoner.

\bibliography{iclr2025_conference}
\bibliographystyle{iclr2025_conference}

\appendix
\section{Appendix}
\section{In-context learning prompts}\label{app:ICL_prompts}
We prepend the following ICL prompts to the test-entries in the experiments in section \ref{sec:experiments}. The \%-character is a comment-character in \texttt{clingo}, and the \#-character denotes the beginning of a new example.
\subsection{Instruction length 1}

\begin{lstlisting}
You are an text to Answer Set Program (ASP) translator. 
Users will ask you questions in English and you will generate a 
ASP program based on the provided  input.
Example 1:
### Instruction:
Generate the Answer Set Program(ASP) for the given input problem.
Each numbered point in the input will have one corresponding ASP
fact as output. Additionally, include some generic ASP rules that
will help the user in solving the problem. 1 X is to the left of K
and is on the same horizontal plane.
2 What is the relation of the agent X to the agent K? 

### Response:
 % we generate one fact for each input sentence: 
% 1 X is to the left of K and is on the same horizontal plane.
left("X", "K").

% 2 What is the relation of the agent X to the agent K?
query("X", "K").

% Now we consider the answer set rules that will help us arrive
at the final answer: 
% assume the 2nd queried object is at location (0,0)
location(Q2, 0, 0) :- query(_, Q2).

% extract answer relation R such that the offset (Ox,Oy) of R is in 
the same direction of (X,Y)
answer(R) :- query(Q1, _), location(Q1, X, Y), offset(R, Ox, Oy),
Ox=-1: X$<$0; Ox=0: X=0; Ox=1: X$>$0; Oy=-1: Y$<$0; Oy=0: Y=0; 
Oy=1: Y$>$0.

% general format translation, which can also be easily done in 
% python script (this is not needed if we directly extract the 
%  general form in the beginning as in bAbI task4)
is(A, top, B) :- top(A, B).
is(A, top, B) :- up(A, B).
is(A, down, B) :- down(A, B).
is(A, left, B) :- left(A, B).
is(A, right, B) :- right(A, B).
is(A, top_left, B) :- top_left(A, B).
is(A, top_right, B) :- top_right(A, B).
is(A, down_left, B) :- down_left(A, B).
is(A, down_right, B) :- down_right(A, B).
is(A, east, B) :- east(A, B).
is(A, west, B) :- west(A, B).
is(A, south, B) :- south(A, B).
is(A, north, B) :- north(A, B).

% synonyms
synonyms(north, northOf; south, southOf; west, westOf; east, 
eastOf; top, northOf; down, southOf; left, westOf; right,
eastOf).
synonyms(A, B) :- synonyms(B, A).
synonyms(A, C) :- synonyms(A, B), synonyms(B, C), A!=C.

% define the offsets of 8 spacial relations
offset(overlap,0,0; top,0,1; down,0,-1; left,-1,0; right,1,0; 
top_left,-1,1; top_right,1,1; down_left,-1,-1; down_right,1,-1).

% derive the kind of spacial relation from synonyms and offset
is(A, R1, B) :- is(A, R2, B), synonyms(R1, R2).
is(A, R1, B) :- is(B, R2, A), offset(R2,X,Y), offset(R1,-X,-Y).

% derive the location of every object
% the search space of X or Y coordinate is within -100 and 100 
(to avoid infinite loop in clingo when data has error)
nums(-100..100).

location(A, Xa, Ya) :- location(B, Xb, Yb), nums(Xa), nums(Ya), 
is(A, Kind, B), offset(Kind, Dx, Dy), Xa-Xb=Dx, Ya-Yb=Dy.

location(B, Xb, Yb) :- location(A, Xa, Ya), nums(Xb), nums(Yb),
is(A, Kind, B), offset(Kind, Dx, Dy), Xa-Xb=Dx, Ya-Yb=Dy.
Example 2: ### Instruction:
Generate the Answer Set Program(ASP) for the given input problem.
Each numbered point in the input will have one corresponding ASP
fact as output. Additionally, include some generic ASP rules
that will help the user in solving the problem. 1 G is at the 6 o
'clock position relative to R.
2 What is the relation of the agent G to the agent R? 

### Response:
% we generate one fact for each input sentence: 
% 1 G is at the 6 o'clock position relative to R.
down("G", "R").

% 2 What is the relation of the agent G to the agent R?
query("G", "R").

% Now we consider the answer set rules that will help us arrive
 at the final answer: 
% assume the 2nd queried object is at location (0,0)
location(Q2, 0, 0) :- query(_, Q2).

% extract answer relation R such that the offset (Ox,Oy) of R is
% in the same direction of (X,Y)
answer(R) :- query(Q1, _), location(Q1, X, Y), offset(R, Ox, Oy),
Ox=-1: X$<$0; Ox=0: X=0; Ox=1: X$>$0; Oy=-1: Y$<$0; Oy=0: Y=0;
Oy=1: Y$>$0.

% general format translation, which can also be easily done in
python script
% (this is not needed if we directly extract the general form in 
the beginning as in bAbI task4)
is(A, top, B) :- top(A, B).
is(A, top, B) :- up(A, B).
is(A, down, B) :- down(A, B).
is(A, left, B) :- left(A, B).
is(A, right:%s/, B) :- right(A, B).
is(A, top_left, B) :- top_left(A, B).
is(A, top_right, B) :- top_right(A, B).
is(A, down_left, B) :- down_left(A, B).
is(A, down_right, B) :- down_right(A, B).
is(A, east, B) :- east(A, B).
is(A, west, B) :- west(A, B).
is(A, south, B) :- south(A, B).
is(A, north, B) :- north(A, B).

% synonyms
synonyms(north, northOf; south, southOf; west, westOf; east, 
eastOf; top, northOf; down, southOf; left, westOf; right,
eastOf).
synonyms(A, B) :- synonyms(B, A).
synonyms(A, C) :- synonyms(A, B), synonyms(B, C), A!=C.

% define the offsets of 8 spacial relations
offset(overlap,0,0; top,0,1; down,0,-1; left,-1,0; right,1,0; 
top_left,-1,1; top_right,1,1; down_left,-1,-1; down_right,1,-1).

% derive the kind of spacial relation from synonyms and offset
is(A, R1, B) :- is(A, R2, B), synonyms(R1, R2).
is(A, R1, B) :- is(B, R2, A), offset(R2,X,Y), offset(R1,-X,-Y).

% derive the location of every object
% the search space of X or Y coordinate is within -100 and 100 
(to avoid infinite loop in clingo when data has error)
nums(-100..100).

location(A, Xa, Ya) :- location(B, Xb, Yb), nums(Xa), nums(Ya), 
is(A, Kind, B), offset(Kind, Dx, Dy), Xa-Xb=Dx, Ya-Yb=Dy.

location(B, Xb, Yb) :- location(A, Xa, Ya), nums(Xb), nums(Yb), 
is(A, Kind, B), offset(Kind, Dx, Dy), Xa-Xb=Dx, Ya-Yb=Dy. 
Example 3:
### Instruction:
\end{lstlisting}

\subsection{Instruction length 2+4}
\begin{lstlisting}
You are an text to Answer Set Program (ASP) translator.Users will 
ask you questions in English and you will generate a ASP program
based on the provided  input.
Example 1:
### Instruction:
Generate the Answer Set Program(ASP) for the given input problem.
Each numbered point in the input will have one corresponding ASP
fact as output. Additionally, include some generic ASP rules that
will help the user in solving the problem. 1 F and T are next to 
each other with F on the left and T on the right.
2 U and F are parallel, and U is on top of F.
3 What is the relation of the agent U to the agent F? 

### Response:
% we generate one fact for each input sentence: 
% 1 F and T are next to each other with F on the left and T on the 
% right.
right("T", "F").

% 2 U and F are parallel, and U is on top of F.
top("U", "F").

% 3 What is the relation of the agent U to the agent F?
query("U", "F").

% Now we consider the answer set rules that will help us arrive at 
% the final answer: 
% assume the 2nd queried object is at location (0,0)
location(Q2, 0, 0) :- query(_, Q2).

% extract answer relation R such that the offset (Ox,Oy) of R is in
% the same direction of (X,Y)
answer(R) :- query(Q1, _), location(Q1, X, Y), offset(R, Ox, Oy),
Ox=-1: X$<$0; Ox=0: X=0; Ox=1: X$>$0; Oy=-1: Y$<$0; Oy=0: Y=0;
Oy=1: Y$>$0.

% general format translation, which can also be easily done in  
%python script (this is not needed if we directly extract the
% general form in the beginning as in bAbI task4)
is(A, top, B) :- top(A, B).
is(A, top, B) :- up(A, B).
is(A, down, B) :- down(A, B).
is(A, left, B) :- left(A, B).
is(A, right, B) :- right(A, B).
is(A, top_left, B) :- top_left(A, B).
is(A, top_right, B) :- top_right(A, B).
is(A, down_left, B) :- down_left(A, B).
is(A, down_right, B) :- down_right(A, B).
is(A, east, B) :- east(A, B).
is(A, west, B) :- west(A, B).
is(A, south, B) :- south(A, B).
is(A, north, B) :- north(A, B).

% synonyms
synonyms(north, northOf; south, southOf; west, westOf; east, 
eastOf; top, northOf; down, southOf; left, westOf; right, 
eastOf).
synonyms(A, B) :- synonyms(B, A).
synonyms(A, C) :- synonyms(A, B), synonyms(B, C), A!=C.

% define the offsets of 8 spacial relations
offset(overlap,0,0; top,0,1; down,0,-1; left,-1,0; right,1,0;
top_left,-1,1; top_right,1,1; down_left,-1,-1; down_right,1,-1).

% derive the kind of spacial relation from synonyms and offset
is(A, R1, B) :- is(A, R2, B), synonyms(R1, R2).
is(A, R1, B) :- is(B, R2, A), offset(R2,X,Y), offset(R1,-X,-Y).

% derive the location of every object
% the search space of X or Y coordinate is within -100 and 100 
% (to avoid infinite loop in clingo when data has error)
nums(-100..100).

location(A, Xa, Ya) :- location(B, Xb, Yb), nums(Xa), nums(Ya), 
is(A, Kind, B), offset(Kind, Dx, Dy), Xa-Xb=Dx, Ya-Yb=Dy.

location(B, Xb, Yb) :- location(A, Xa, Ya), nums(Xb), nums(Yb), 
is(A, Kind, B), offset(Kind, Dx, Dy), Xa-Xb=Dx, Ya-Yb=Dy.
Example 2:
### Instruction:
Generate the Answer Set Program(ASP) for the given input problem.
Each numbered point in the input will have one corresponding ASP
fact as output. Additionally, include some generic ASP rules that 
will help the user in solving the problem. 
1 C and M are both there with the object C above the object M.
2 Z is at the bottom and Y is on the top.
3 Z is at a 45 degree angle to M, in the upper lefthand corner.
4 Y is placed at the lower left of G.
5 What is the relation of the agent Z to the agent C? 
### Response:
% we generate one fact for each input sentence: 
% 1 C and M are both there with the object C above the object M.
top("C", "M").

% 2 Z is at the bottom and Y is on the top.
down("Z", "Y").

% 3 Z is at a 45 degree angle to M, in the upper lefthand corner.
top_left("Z", "M").

% 4 Y is placed at the lower left of G.
down_left("Y", "G").

% 5 What is the relation of the agent Z to the agent C?
query("Z", "C").

% Now we consider the answer set rules that will help us arrive
% at the final answer: 
% assume the 2nd queried object is at location (0,0)
location(Q2, 0, 0) :- query(_, Q2).

% extract answer relation R such that the offset (Ox,Oy) of R is
% in the same direction of (X,Y)
answer(R) :- query(Q1, _), location(Q1, X, Y), offset(R, Ox, Oy), 
Ox=-1: X$<$0; Ox=0: X=0; Ox=1: X$>$0; Oy=-1: Y$<$0; Oy=0: Y=0;
Oy=1: Y$>$0.

% general format translation, which can also be easily done in  
% python script (this is not needed if we directly extract the
% general form in the beginning as in bAbI task4)
is(A, top, B) :- top(A, B).
is(A, top, B) :- up(A, B).
is(A, down, B) :- down(A, B).
is(A, left, B) :- left(A, B).
is(A, right, B) :- right(A, B).
is(A, top_left, B) :- top_left(A, B).
is(A, top_right, B) :- top_right(A, B).
is(A, down_left, B) :- down_left(A, B).
is(A, down_right, B) :- down_right(A, B).
is(A, east, B) :- east(A, B).
is(A, west, B) :- west(A, B).
is(A, south, B) :- south(A, B).
is(A, north, B) :- north(A, B).

% synonyms
synonyms(north, northOf; south, southOf; west, westOf; east, 
eastOf; top, northOf; down, southOf; left, westOf; right,
eastOf).
synonyms(A, B) :- synonyms(B, A).
synonyms(A, C) :- synonyms(A, B), synonyms(B, C), A!=C.

% define the offsets of 8 spacial relations
offset(overlap,0,0; top,0,1; down,0,-1; left,-1,0; right,1,0; 
top_left,-1,1; top_right,1,1; down_left,-1,-1; down_right,1,-1).

% derive the kind of spacial relation from synonyms and offset
is(A, R1, B) :- is(A, R2, B), synonyms(R1, R2).
is(A, R1, B) :- is(B, R2, A), offset(R2,X,Y), offset(R1,-X,-Y).

% derive the location of every object
% the search space of X or Y coordinate is within -100 and 100 
% (to avoid infinite loop in clingo when data has error)
nums(-100..100).

location(A, Xa, Ya) :- location(B, Xb, Yb), nums(Xa), nums(Ya), 
is(A, Kind, B), offset(Kind, Dx, Dy), Xa-Xb=Dx, Ya-Yb=Dy.

location(B, Xb, Yb) :- location(A, Xa, Ya), nums(Xb), nums(Yb), 
is(A, Kind, B), offset(Kind, Dx, Dy), Xa-Xb=Dx, Ya-Yb=Dy. 
Example 3:
### Instruction:
\end{lstlisting}

\end{document}